\documentclass{article} 
\usepackage{iclr2021_conference,times}

\usepackage{amsmath,amsfonts,bm}

\def\Figref#1{Figure~\ref{#1}}

\def\Secref#1{Section~\ref{#1}}

\def\eqref#1{equation~\ref{#1}}

\def\1{\bm{1}}

\DeclareMathAlphabet{\mathsfit}{\encodingdefault}{\sfdefault}{m}{sl}
\SetMathAlphabet{\mathsfit}{bold}{\encodingdefault}{\sfdefault}{bx}{n}

\usepackage{graphicx}
\usepackage{hyperref}
\usepackage{url}
\usepackage{booktabs}
\usepackage{listings}
\usepackage{xcolor}
\usepackage{wrapfig}
\usepackage{enumitem}

\usepackage{algorithm}
\usepackage{algorithmicx}
\usepackage{algpseudocode}
\usepackage{svg}

\newcommand{\hide}[1]{}
\newcommand{\vpara}[1]{\vspace{0.07in}\noindent\textbf{#1.}}

\newcommand\fmoe{\textit{FastMoE}\ }
\newcommand\Fmoe{FastMoE}

\title{\Fmoe: A Fast Mixture-of-Expert Training System}

\hide{
\title{\fmoe: A Fast Mixture-of-Experts Training System for High Efficiency}
\author{Jiaao He \\
\texttt{hja20@mails.tsinghua.edu.cn}}
\institute{Tsinghua University}
}
\author{Jiaao He$^{\dag\ddag}$, Jiezhong Qiu$^{\dag\ddag}$, Aohan Zeng$^{\dag\ddag}$, Zhilin Yang$^{\ddag\sharp}$, Jidong Zhai$^{\dag\ddag}$, Jie Tang$^{\dag\ddag}$ \\
$^\dag$~Tsinghua University 
$^\ddag$~Beijing Academy of Artificial Intelligence~(BAAI) 
$^{\sharp}$~Recurrent AI\\
\texttt{\{hja20,qiujz16,zah18\}@mails.tsinghua.edu.cn};\\
\texttt{kimi\_yang@rcrai.com};
\texttt{\{zhaijidong, jietang\}@tsinghua.edu.cn}}

\iclrfinalcopy 

\begin{document}

\maketitle
\begin{abstract}
Mixture-of-Expert~(MoE) presents a strong potential in enlarging the size of language model to trillions of parameters.
However, training trillion-scale MoE requires 
algorithm and system co-design for a well-tuned high performance distributed training system.
Unfortunately, 
the only existing platform that meets the requirements
strongly depends on Google's  hardware~(TPU) and software~(Mesh Tensorflow) stack, and is 
  not open and available to the public, especially GPU and PyTorch communities.

In this paper, we present \fmoe, a distributed MoE training system based on PyTorch with common accelerators.
The system provides a hierarchical interface for both flexible model design and easy adaption to different applications, such as Transformer-XL and Megatron-LM.
Different from direct implementation of MoE models using PyTorch, the training speed is highly optimized in \fmoe by sophisticated high-performance acceleration skills.
The system supports placing different experts on multiple GPUs across multiple nodes,
enabling enlarging the number of experts linearly against the number of GPUs.
The source of \fmoe is available at \url{https://github.com/laekov/fastmoe} under Apache-2 license.

\end{abstract}

\section{Introduction}
\hide{
Checklist:
	1. What is the problem?
	2. Why is it important?
	3. Why is it hard?
	4. Why hasn't it been solved already? (How does our solution advance the state of the art exactly?)
    5. What is our solution?
}

Recent emergence of large-scale language models, examplified by BERT~\citep{devlin2018bert}, GPT-2/-3~\citep{radford2019language, brown2020language}, XLNet~\citep{yang2019xlnet}, RoBERTa~\citep{liu2019roberta}, T5~\citep{raffel2020exploring}, GShard~\citep{chen2020gshard} and Switch Transformer~\citep{fedus2021switch}, has drastically reshaped the landscape of the natural language processing research, reestablishing the new state-of-the-art baselines in various benchmarks such as GLUE~\citep{wang2018glue} and SuperGLUE~\citep{wang2019superglue}.

Among many possible solutions, scaling model size has been proved to be one of the simplest and most effective way~\citep{kaplan2020scaling} toward more powerful models.
\hide{
In fact, the sizes of these models have increased by four orders of magnitude in past years --- from BERT's 330 million parameters to Switch Transformer's 1571 billion parameters.
}
From BERT~\citep{devlin2018bert} with $340$~million parameters, 
to T5~\citep{raffel2020exploring} with $11$~billion parameters,
to GPT-3~\citep{brown2020language} with $175$~billion parameters,
the model size is enlarged by $500\times$ in merely two years.
%
More recently, GShard~\citep{chen2020gshard} scales to a record number of $600$ billion parameters, which is quickly broken by Switch Transformer~\citep{fedus2021switch} with $1.6$ trillion parameters.
The main contributor toward the huge model size of GShard and Switch Transformer is a novel neural network architecture named  mixture of experts~(MoE)~\citep{shazeer2017outrageously}.

\begin{figure}[ht]
    \centering
    \includegraphics[width=.9\textwidth]{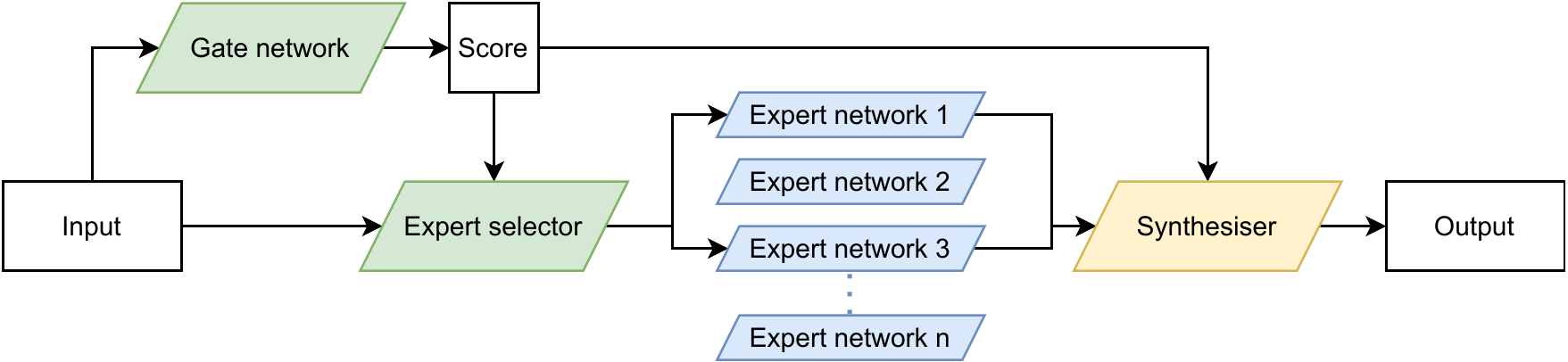}
    \caption{An illustrative example of an MoE layer. \small In this example, expert 1 and expert 3 are selected by the gate for computation.}
    \label{fig:moe}
\end{figure}

\hide{
\begin{wrapfigure}{r}{0.3\textwidth}
\centering
\vspace{-0.2in}
\includegraphics[width=0.3\textwidth]{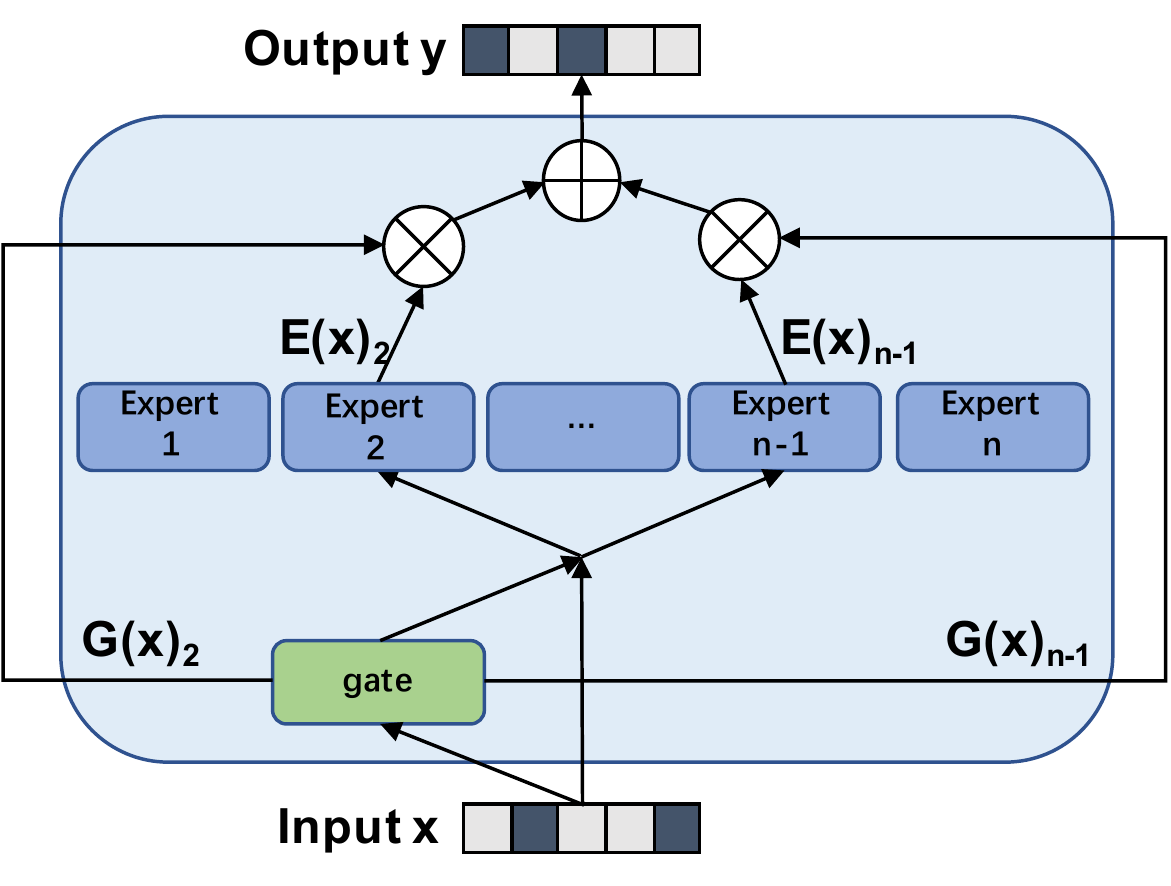}
 \vspace{-0.2in}
  \label{fig:moe_demo}
\end{wrapfigure}
}

\hide{
\begin{figure}[ht]
\centering
\includegraphics[width=0.6\textwidth]{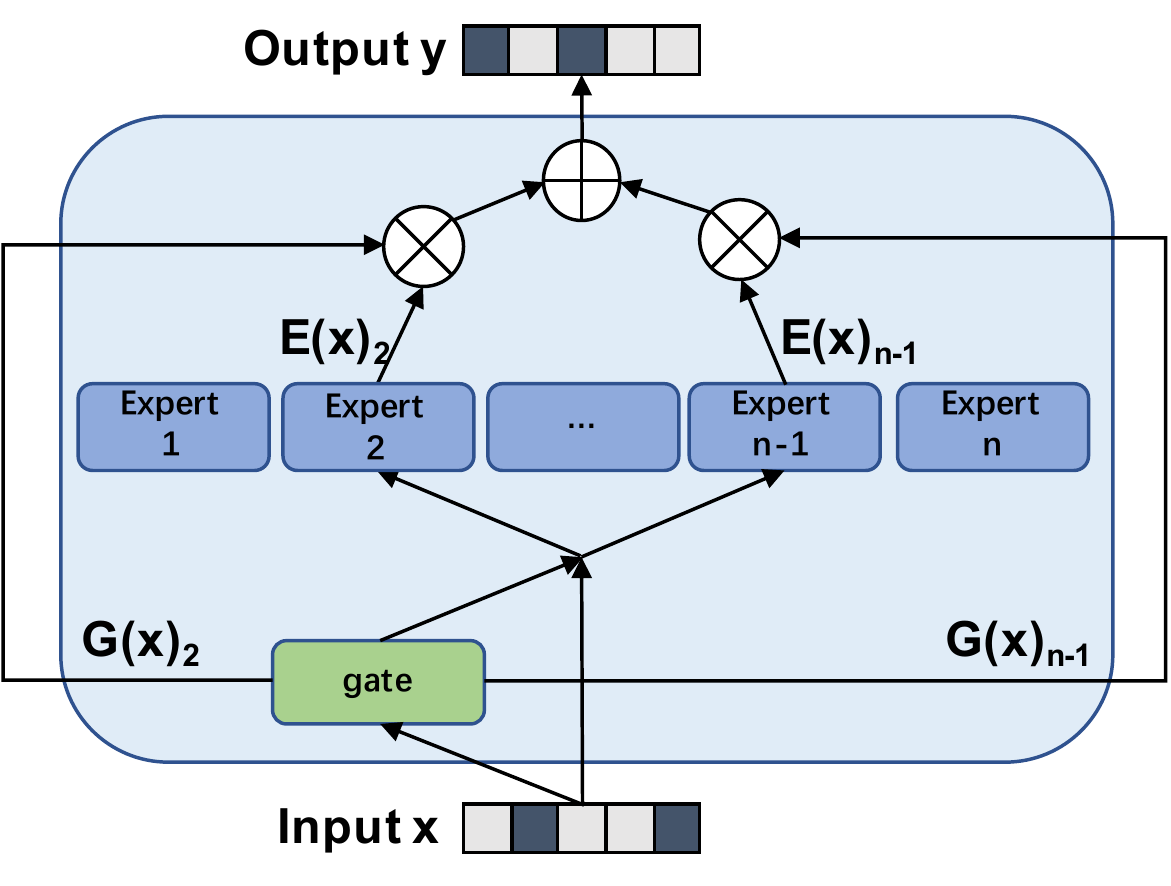}
  \label{fig:moe_demo}
  \caption{An example of an MoE layer}
\end{figure}
}
An MoE layer~(an illustrative example can be found in \Figref{fig:moe}) consists of a gate and a pool of experts.
For each input, only a tiny minority of experts are chosen by the gate for computation.
The special architecture of MoE 
is a double-edge sword for large-scale distributed training. On the one hand, due to its sparse activation of experts, MoE is able to enlarge the model size by orders of magnitude without significantly increasing the amount of computation~(FLOPs).
On the other hand, when scaling up to thounds of experts, the imbalanced all-to-all communication pattern of MoE brings new challenges to the co-design of algorithm and system.
Therefore, MoE can not be directly supported by traditional deep learning libraries such as PyTorch~\citep{NEURIPS2019_9015} and TensorFlow~\citep{abadi2016tensorflow}.

Due to the challenge raised by the new model architecture, both the research community and the industry need 
an MoE implementation that support large-scale distributed training.
However, 
despite the existence of some naive single-GPU implementation in PyTorch~\citep{rau2019moe}, 
the only present system 
that supports scalable MoE training is based on Google's private hardware and software stack --- TPU~\citep{jouppi2017datacenter} and Mesh TensorFlow~\citep{shazeer2018mesh}.
Thus, there is urgent need to develope an MoE system on publicly available hardware~(e.g., GPU) and platforms~(e.g., PyTorch~\citep{NEURIPS2019_9015}).

\hide{
The academia has been actively looking into MoE models.
Unfortunately, a system based on TPU and mesh-tensorflow~\citep{shazeer2018mesh} is the only present implementation that supports large-scale training of  MoE models.
Open source versions of PyTorch implemented MoE models, such as \cite{rau2019moe}, are developed.
However, the open source models are of fixed structure, and they do not have distributed training support, 
which means only one GPU can be used to train the model.
Therefore, researchers cannot train large distributed MoE models with publicly available resources.}

Motivated by the desire to obtain easy-to-use, flexible, efficient, scalable, and open-source solution 
to large-scale MoE training, we release \fmoe with the following design goals:
\begin{itemize}[topsep=1pt,itemsep=0pt,parsep=0pt,leftmargin=15pt]
    \item \textbf{Easy-to-use}: provide a user-friendly interface to define an MoE layer, and seamless support for popular language model training system, Megatron-LM~\citep{shoeybi2019megatron}.
    \item \textbf{Flexible}: make it easy for users to customize gate networks and expert networks.
    \item \textbf{Efficient}: integrate a highly optimized feadforward~(FFN) layer for Transformer.
    \item \textbf{Scalable}: support scaling up the size of MoE models by training across multiple GPUs on multiple nodes.
\end{itemize}

\hide{
In this paper, an open-source system, \fmoe, is introduced to enable research on large MoE models.
\fmoe is designed for usability and flexibility.
For users who are willing to explore different models to extend with MoE, the core level of \fmoe supports very flexible expert defination.
For users who want to use \fmoe in their own transformer model, \fmoe provides a highly optimized FFN layer in transformers with distributed support.
For users using predefined models, such as Megatron-LM \citep{shoeybi2019megatron}, \fmoe contains adapters for one-key \textit{fmoefying} the model, 
i.e. replicating some layers in the model to make it an MoE model that can be computed using \fmoe.
}

Different from previous single-GPU PyTorch implementation~\citep{rau2019moe}, \fmoe concentrates on efficiency and scalability.
Dedicated CUDA kernels are included in \fmoe for high performance with specialized optimizations.
\fmoe is able to run across multiple GPUs on multiple nodes using NCCL~\citep{jeaugey2017nccl}.
The details of communication is hidden from model developers by \fmoe.
The model-parallel method of \fmoe allows distributing experts across different GPUs,
while other parts of the model remain parallelized by batch dimension~(data parallel) or tensor dimension~(model parallel).
Chances are that the model size, proportional to the number of experts, can scale up with the number of GPUs used for training,
being the key to train trillion-scale models.

In our experiment, we observe that \fmoe is faster than a baseline \citep{rau2019moe} implemented by pure PyTorch API on a single GPU.
\fmoe also shows reasonable scalability when running across nodes on a cluster connected by Infiniband network.
We train a real GPT model with $96$ experts per layer using distributed \fmoe with promising end-to-end training speed.
Compared to a non-MoE model of the same amount of computation,
its performance benefits from the enlarged model size that the MoE architecture.

This paper is organized as follows.
\Secref{sec:moe} introduces the background of MoE and compares existing systems.
\Secref{sec:sys} presents the \fmoe system in detail.  
\Secref{sec:rhp} introduces the challenges of achieving high-performance and the \fmoe's solutions.
\Secref{sec:evl} shows results of experiments that demonstrate the efficiency of \fmoe and the performance gain of an MoE model using \fmoe in training.
\Secref{sec:sum} summarizes the paper and indicates directions our future work.


\section{Mixture-of-Experts~(MoE)}
\label{sec:moe}
In this section, we review the architecture of MoE
and current systems for training MoE.

\subsection{MoE: Model Structure}

Mixture-of-Expert is short for Sparsely-Gated 
Mixture-of-Experts layers proposed by \cite{shazeer2017outrageously}.
An MoE layer consists of multiple experts, each can be an arbitrary neural network.
The only constraint of the experts is that they should take the same input, and give output in the same vector space.
\Figref{fig:moe} shows a detailed example of an MoE layer.
A special neural network, namely \textit{the gate network}, is introduced to score each expert over a given input.
According to the scores, selection of experts is made by a policy, which may vary from model to model.
Then, the selected experts, e.g. experts $1$ and $3$ in the example, are activated to process the input sample.
The outputs of the experts, together with the score, are combined into the final output using a certain algorithm.

A popular way of the expert selection is to select the experts with top $k$ highest score.
In synthesis process, the score is used as the weight for the output of the experts to be added into the overall output.
This enables training the gate network, as the gradient can be propagated by the score.
Algorithm \ref{algo:moe} formalizes the method above.

\begin{algorithm}[ht]
    \caption{Forward computation of an MoE layer with top-$k$ gating.}
    \label{algo:moe}
    \begin{algorithmic}[1]
    \Require{A pool of $n$ experts: $\{E_1, E_2, \cdots, E_n\}$}
    \Require{Gate $G$}
    \Require{The number of experts $k$ to be selected}
    \Function {MoE}{$x$}
        \State $score \Leftarrow G(x)$
        \State $indices \Leftarrow ArgMax_k(score)$
        \State $y \Leftarrow$ zero tensor like $x$
        \For{each index $i \in indices$}
            \State $x_i \Leftarrow E_i(x)$
            \State $y \Leftarrow score_i * x_i + y$
        \EndFor
        \State \Return $y$
    \EndFunction
    \end{algorithmic}
\end{algorithm}

\subsection{Current Systems for MoE Training}

The GShard system~\citep{chen2020gshard} implements a distributed version of the MoE model.
It trains a language model on up to $2048$ TPUs, with $1$ expert per layer placed on each TPU.
As a result, the MoE layers contain $2048 \times$ more parameters than a non-MoE layer.
In Switch Transformer~\citep{fedus2021switch}, the model is further enlarged to $1.6$ trillion,
showing strong ability for the system to support training models in large scale.
Unfortunately, this system is not publicly available yet. 
It is strongly binded with the TPU cluster, which makes it hard to reproduce the experiments on commodity devices.
Additionally, the design of GShard lacks flexibility to use different number and size of experts with different replication strategy.

In Tensor2tensor~\citep{tensor2tensor}, a MoE Transformer model is provided.
However, this implementation uses Mesh TensorFlow \citep{shazeer2018mesh}, which does not support GPUs very well.
To implement a FFN in Transformer, it takes more than $100$ lines of code in TensorFlow with complicated \texttt{einsum} operators,
making it burdening for developers to understand the structure and explore other model structures based on the code.

PyTorch~\citep{NEURIPS2019_9015}, as a popular deep learning framework among researchers, provides more straightforward coding style and flexibility against TensorFlow.
Efforts are made to train MoE models with PyTorch \citep{rau2019moe}.
However, as the PyTorch community lacks multi-dimension parallel training tools, none of the PyTorch-based implementations support training on multiple GPUs.
As the ultimate target of adopting MoE is training even larger models, the PyTorch-based implementations fails to be a candidate.

\section{\fmoe: System Design} 

\label{sec:sys}

In this section,
we introduce our design of \fmoe with distributed training support.

\subsection{A Flexible System for Diverse Model Explorers}

\vpara{The Backbone to Run Arbitrary Expert Networks}
\fmoe supports using arbitrary network as the expert.
The \texttt{FMoE} interface of \fmoe takes any neural network module constructor as input,
and replicates the module for multiple times as the expert instances.
The expert is defined to take a batch of aligned contiguous input features, and the output should be in the same batch order.
Therefore, the expert module implementation is decoupled from the MoE architecture so that developers can focus on the design of their own expert network.

For even stronger flexibility, the \texttt{FMoE} class contains a member function \texttt{expert\_fn}, 
where the expert modules are used to conduct the forward computation.
This function can be overloaded for further customized MoE behavior.
For example, in the \texttt{FMoETransformerMLP} network, which will be mentioned later in this section.
the list of experts is replaced by a specially optimized module that applies the experts in parallel to extremely lower the latency.

Moreover, \fmoe supports placing multiple experts together on the same worker, 
enabling more flexible configuration space of the number of experts~(i.e., the number of experts does not have to be equal to the number of data parallels),
which is different from the the design of  GShard.

\vpara{A Highly-optimized FFN for Transformer}
To better support training Transformer with MoE,
\fmoe  provides a standard and high-performance FFN implementation~(\texttt{FMoETransformerMLP}). 
The detailed optimization strategy is hidden from the developers.

In particular, when placing multiple experts on the same worker, a naive implementation is to loop over these experts and conduct forward in sequence. However, for certain types of expert networks, it is possible to explore the potential speedup brought by parallel execution.
In \fmoe, we mainly optimize the parallel execution of fully-connected layers by 
a dedicated \texttt{FMoELinear} module. 
Instead of computing the expert modules sequentially, the specially optimized expert module maintains a pool of available hardware resources,
and applies the expert computation in parallel.

\hide{
A number of fully-connected layers, as the expert of the Transformer model, make up a large portion of the computation time.
As multiple experts are located on the same worker, 
it is intuitive to perform the computation of different experts simultaneously to explore the potential speedup brought by parallel execution.
}

\vpara{Plugin-style Support for PyTorch and Megatron-LM}
\hide{
The \fmoe system is based on PyTorch \citep{NEURIPS2019_9015}, a popular neural network training framework.
However, as PyTorch only provides lower level tensor algebra support with automatic gradient computation,
much additional effort, e.g. input preprocessing, building the model and sophisticated parallelization, is needed.
Therefore, various models with well-organized infrastructure are wrapped into training applications.
For example, Transformer-XL \citep{transformer-xl} provides data downloading and preprocessing scripts that users can conveniently start training on given datasets.
Megatron-LM \citep{shoeybi2019megatron} implements multiple parallel strategies, which are very user-friendly.
}
The flexibility of \fmoe allows convenient adaption to existing training applications.
Take Megatron-LM~\citep{shoeybi2019megatron} as an example,
a plugin-style module is integrated in \fmoe to quickly replace the FFNs in the original Megatron-LM model with MoE networks.
As shown in listing \ref{lst:megatron}, the transformation can be achieved by only 2 lines of code.

\begin{lstlisting}[language=Python, caption=Sample code to use \fmoe in Megatron-LM, label=lst:megatron]
from fmoe.megatron import fmoefy
model = fmoefy(model, num_experts=<number of experts per worker>)
\end{lstlisting}

The \texttt{fmoefy} function can find the FFNs in the Transformer layers. 
Then, an MoE network using \fmoe is created, which is a module that wraps up the \texttt{FMoETransformerMLP} module for interface-level compatibility.

\subsection{Enlarging the Model Capacity Distributedly}

\vpara{The Model Parallel Method of \textit{FastMoE}}
As one of the most effective way to enlarge the model capacity,
the ability to accommodate a large expert population and train them in parallel is demanded in many MoE models.
It is hard for the model developers to handle the complicated data transfer among GPUs and even across nodes.
Achieving high training performance and good hardware resource utilization requires expertise in computer architecture and parallel programming,
which is beyond the technique stack of common model developers.

\fmoe supports distributing experts across multiple workers on multiple nodes,
which is called \textit{the model parallel method in \fmoe}.
The detail of input data exchange is hidden within the \texttt{FMoE} interface.
For model developers, they only need to write code for a single expert,
and each expert is given all the input data gathered from all workers by \fmoe.
As a result, the model developers do not have to consider the implementation details about cross-worker communication.

In the design of \fmoe, when the feature to distribute expert across workers is enabled, 
extra communication operations are included in the forward and backward computation.
To better identify the operations, we call them global data exchange operations, 
in contrast to the local data shuffle process, which will be mentioned in section \ref{sec:rhp}.

A major challenge in the distributed context is that the total number of input samples assigned to all experts on a worker may vary a lot.
It is impossible to have the number of incoming samples before the gate output is available.
However, allocating the buffer to place the input samples is dependent on the number.
Therefore, before actual exchange of input samples between workers happens after exchanging the amount information between workers,
and allocating memory according to the inspection of the expert count information.

\begin{figure}[ht]
    \centering
    \includegraphics[width=\linewidth]{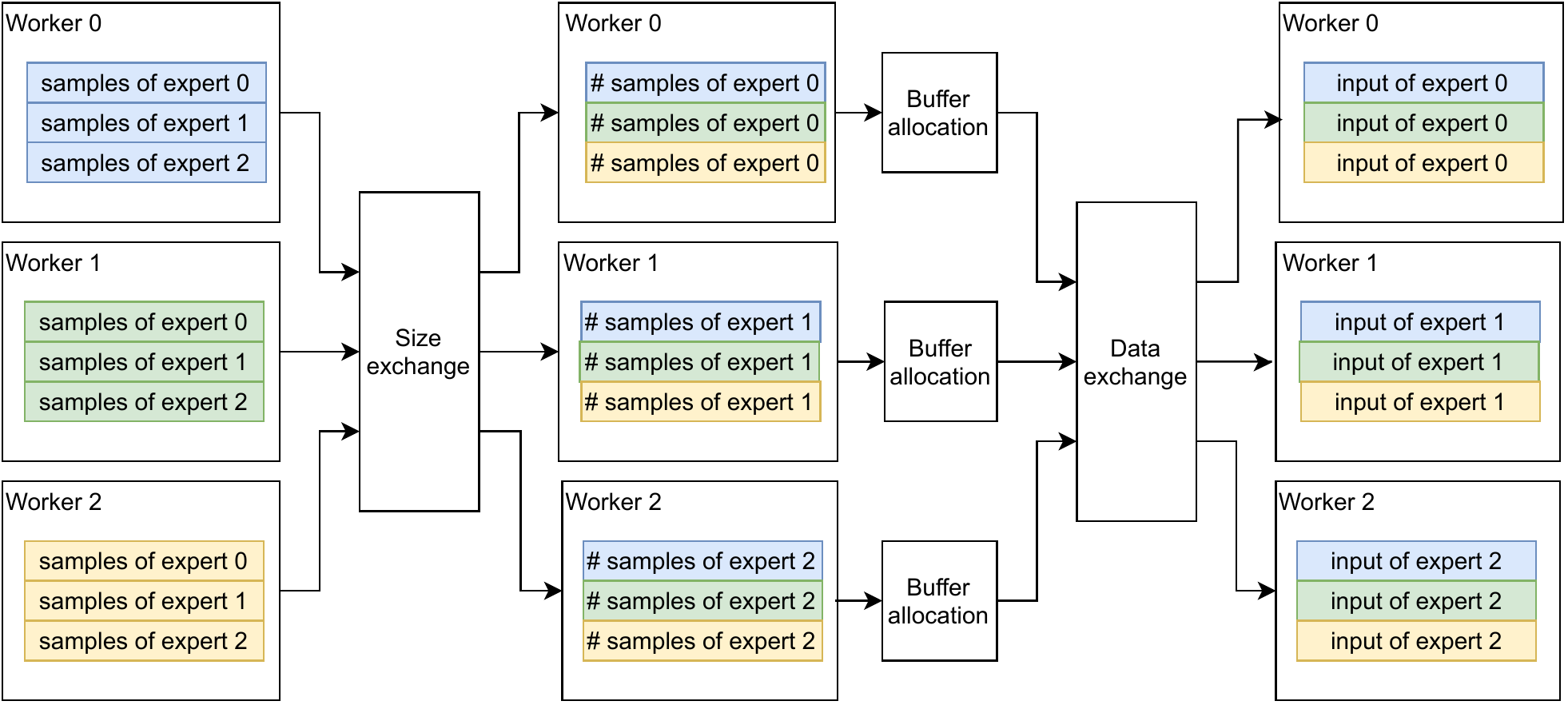}
    \caption{An example of the global operations.}
    \label{fig:glbsc}
\end{figure}

An example of the global operations in \fmoe is shown in figure \ref{fig:glbsc}.
The workers first count the number of samples assigned to each expert on each worker.
Then, they exchange the size of expert inputs, 
so that all workers get the number of incoming input samples, and where they are from.
After the offsets of each receiving buffer is calculated, the workers start exchanging data directly. 
It is worth nothing that the statistics of the incoming and outgoing samples can be reused through the whole process of a training iteration.

\vpara{Heterogeneity-aware Synchronization Module}
Heterogeneity is introduced as different parts of the network may be replicated across different groups of workers.
It is a challenge that the distributed module has to identify whether the gradient of a parameter should be synchronized,
and with whom it is synchronized.
\fmoe introduces the \textit{data parallel communication group} tag on each parameter to address the issue.

The tag can be one of \texttt{world}, \texttt{data parallel} or \texttt{none},
which respectively indicates that the gradient should be synchronized with (1) all other workers,
(2) the workers in a data-parallel group that is orthogonal to the model-parallel group, or (3) no worker.
For example, the gate network is replicated across all workers, regardless of model parallel settings.
The attention layer may be divided into model-parallel sub-layers, so its tag is \texttt{data parallel}.
Each worker serves several unique expert networks, whose tag is \texttt{none}.
A customized data parallel module instead of PyTorch's original distributed data parallel module is provided by \fmoe, which can identify the tags and perform  correct synchronization.

\section{Optimizations to Achieve High-performance}

\label{sec:rhp}

The performance of MoE computation on a single node is significant,
as it determines the theoretical upper bound of the system scaling up to any scale.

The most intuitive way to compute an MoE layer is slicing the input batch into samples, and compute sample by sample.
After that, output features are stacked in the original order.
However, it is observed that implementing an MoE model using simple PyTorch operators can hardly achieve high performance.
Less than $5\%$ the peak performance of GPUs can be achieved.

\begin{figure}[ht]
\centering
\includegraphics[width=.6\textwidth]{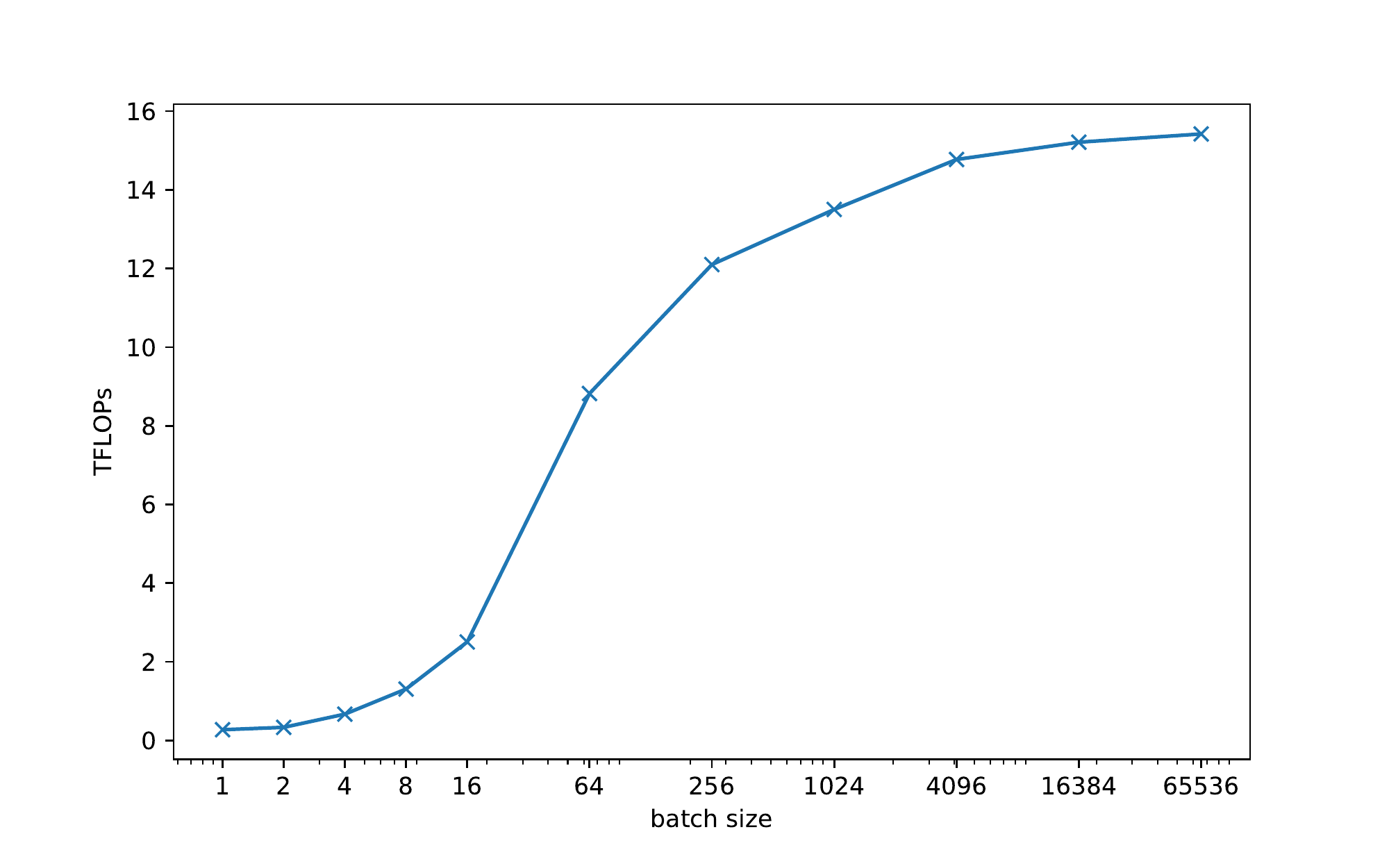}
\caption{\texttt{GeMM} performance of different problem sizes using \texttt{cuBLAS} on NVIDIA V100.}
\vspace{-0.1in}
\label{fig:gemm}
\end{figure}

Without loss of generality, we assume the expert network is an FFN. Note that the major operator within an FFN is from fully-connected layers, which consist of several \texttt{GeMM} operators.
When the batch is split up into single samples, the \texttt{GeMM} operation is degraded into \texttt{GeMV}.
Figure \ref{fig:gemm} shows the float-point computation throughput of an example fully-connected layer using different batch size.
Given that in modern heterogeneous compute devices, 
matrix multiplication operators are fine tuned with sophisticated tiling techniques applied on all dimensions,
it is not surprising that the throughput can only approach the theoretical peak when the batch size is large enough.
This leads to the principle that to achieve high performance in MoE computation, 
the samples should be batched to fully utilize the hardware resources.

\fmoe batches all input samples to the same expert together.
Due to the limit of data representation, \fmoe performs memory movement with a specially developed CUDA kernel to reduce overhead.
Given the index of gate that each sample is going to,
the process to put all input samples to the same gate in a contiguous memory space is called \texttt{scatter}.
However, in other parts of the neural network, the batch may have to be organized as its original order, e.g., the attention layer in Transformer.
A reverse operation is performed after the experts output to another contiguous memory space,
i.e. place the scattered feature vectors back to their original order according to the gate indices.
This process is denoted as \texttt{gather} in \fmoe.

\begin{figure}[ht]
    \centering
    \includegraphics[width=\linewidth]{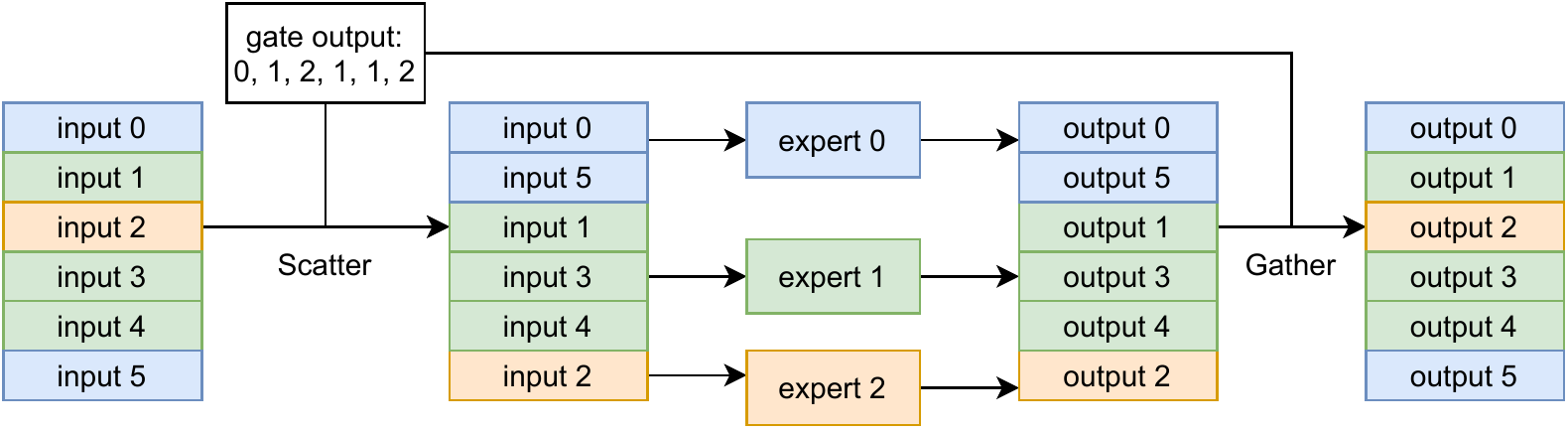}
    \caption{An example of the reordered computation of an MoE layer}
    \label{fig:reorder}
\end{figure}

The reordering computation process is shown as \Figref{fig:reorder}.
When the assignment from input samples to experts is balance enough, 
each expert is expected to have a relatively large input batch size that can reach satisfying hardware utilization according to \Figref{fig:gemm}.
However, load imbalance always occurs because of the nature of random sampling of the input training data.
It is highly possible that one expert receives very few input samples during millions training iterations.
Additionally, as multiple experts are placed on one worker, local batch sizes of the experts are, on average, statistically lower than that in data parallel.
\fmoe uses a customized stream manager to simultaneously execute the computation of multiple experts to extract the potential throughput gain.

\section{Evaluation}
\label{sec:evl}

In this section, the training speed of \fmoe is compared with another PyTorch MoE implementation~\citep{rau2019moe} on a single GPU.
We also report the scalability of \fmoe when distributed training.
To the best of our knowledge, \fmoe is the only  PyTorch-based MoE system can run across different nodes and GPUs.
We also show the end-to-end performance of an MoE Transformer model trained using \fmoe.

\subsection{Experiment Setup}

We use the following notations to characterize the computation task:
$n_e$ experts are placed on each GPU.
Each expert applies two linear layers of sizes $d_m \times d_h$ and $d_h \times d_m$ respectively.
The input contains $n_b$ samples.
The gate module scores the fitness of each sample to be processed by each expert.
For each input sample, the experts of top $k$ highest score are selected to process the sample.

Additionally, several warm-up rounds are performed, which perform the same computation but are not counted in the results.
For each experiment, the task is executed $16$ times, and the average time is used to calculate the performance.
The standard deviation values of the execution time are also inspected.
All of them are negligible.

\subsection{Training Speed on a Single GPU}

The performance of the \texttt{FMoETransformerMLP} is tested, which completes similar task with \texttt{MoE} module in the baseline \citep{rau2019moe},
on a NVIDIA TESLA V100 PCIe GPU.
The baseline is implemented by pure PyTorch API with hard-coded model structure.
For fairness of the comparison, both modules uses a randomly initialized matrix as the weight of the gate network, which consists of one fully-connected layer.
The experts also perform the same computation.

\begin{figure}[ht]
    \centering
    \includegraphics[width=0.8\textwidth]{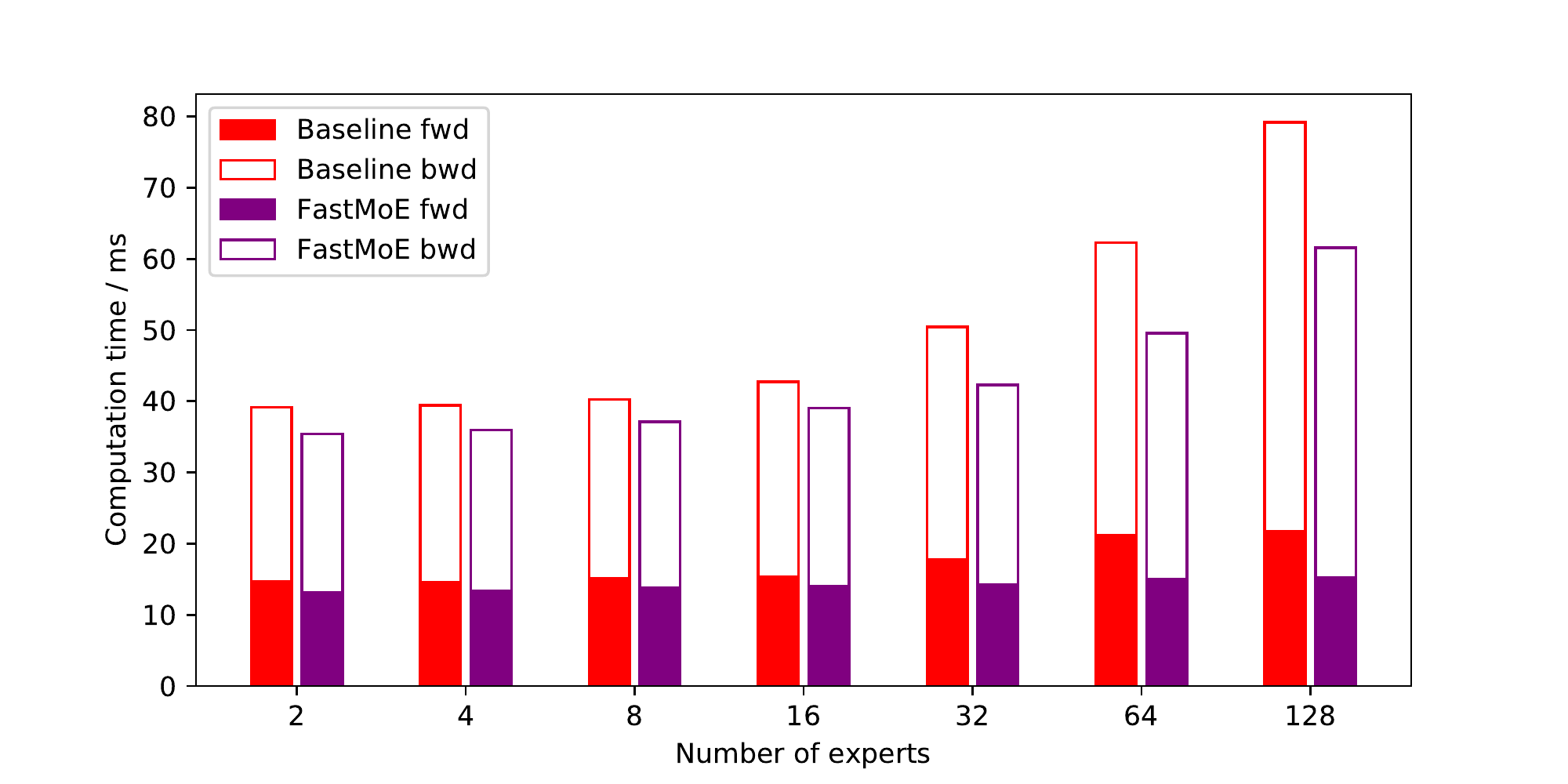}\\
    \footnotesize The latency is tested with $n_b=4096, d_m=1024, d_h=4096, k=2$.
    \caption{Computation time comparison between \fmoe and the baseline implementation.}
    \label{fig:sglres}
\end{figure}

As \Figref{fig:sglres} shows, the baseline implementation is constantly slower than \fmoe.
As the number of experts grows, the baseline spends much more time in the forward computation, 
while the latency of \fmoe remains stable, thanks to its customized stream manager mentioned in \Secref{sec:rhp}.
Considering that \fmoe is targeted on training, the backward time is stacked over the forward time.
We observed that \fmoe outperforms the baseline in the overall time spent in each iteration.

\subsection{Cross-GPU and Cross-node Scalability}

To examine the performance of \fmoe expanding on multiple GPUs across nodes,
we conduct an experiment on a cluster of $8$ nodes, with $1$ NVIDIA Tesla V100 GPU on each node.
The cluster is interconnected via an Infiniband EDR switch and $8$ HCA cards.
The FLOPs of the matrix multiplication operations is calculated to represent the training throughput.

\begin{figure}[ht]
    \centering
    \includegraphics[width=\linewidth]{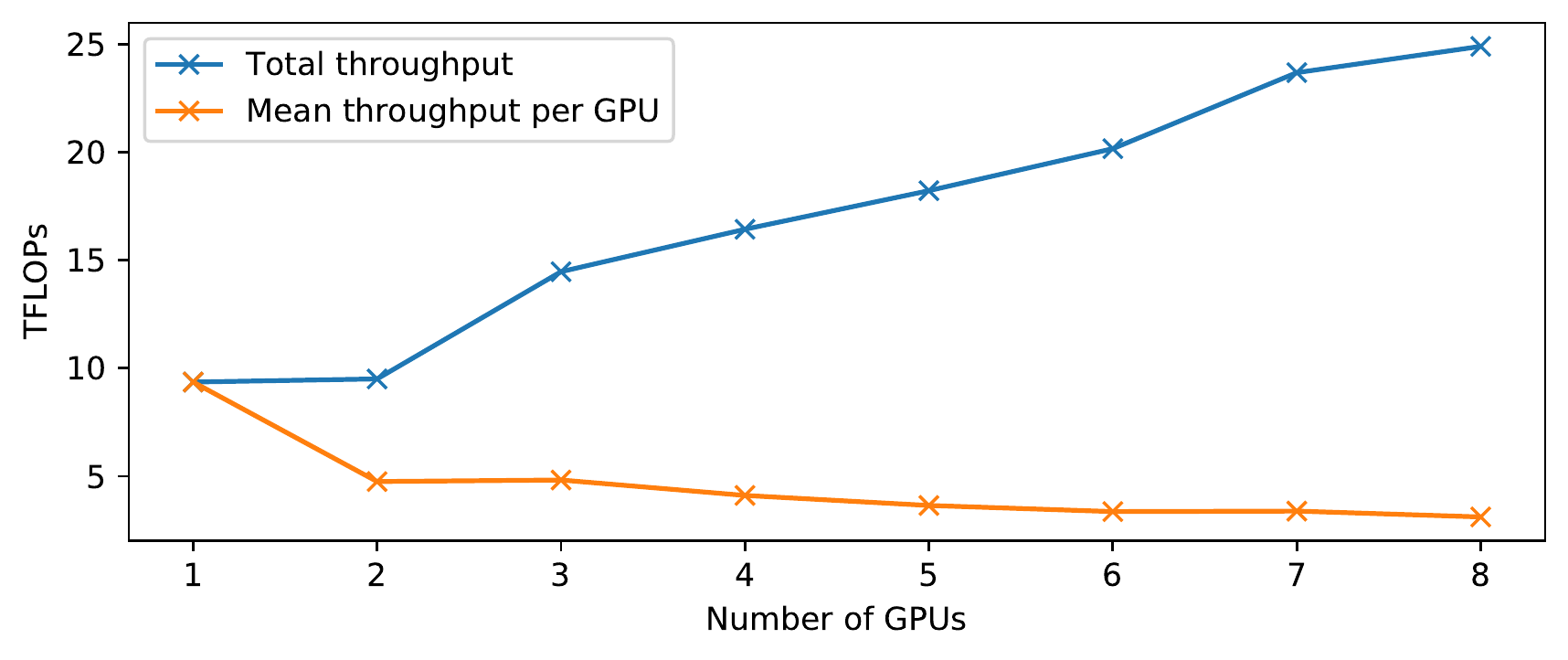}
    \footnotesize The throughput is tested with $n_e=4, n_b=4096, d_m=1024, d_h=4096, k=2$.
    \caption{Scalability of \fmoe across multiple GPUs on multiple nodes}
    \label{fig:mulres}
\end{figure}

According to the result shown in \Figref{fig:mulres}, \fmoe shows scalability across nodes. 
The overall throughput increases from $10$ \texttt{TFLOPs} to $25$ \texttt{TFLOPs}, as the number of GPUs increases from $2$ to $8$, sub-linearly scaling up.
We observe that when expanding to $2$ GPUs, the performance is half of that on a single GPU, which suggests that \fmoe is bounded by communication.
When more GPUs are used for computation, more experts are introduced, and the granularity of exchanging input samples becomes smaller, 
lowering the efficiency in data transfer over the network.

As a conclusion, the scalability of \fmoe can support training large MoE model using multiple GPUs across multiple nodes with performance gain.
However, space is left for further optimization on throughput.

\subsection{End-to-end Performance Gain Using \fmoe}

We test the end-to-end performance gain using \fmoe by training a 12-layer GPT model on $8$ GPUs using Megatron-LM~\citep{shoeybi2019megatron}.
As mentioned in  \Secref{sec:sys}, the Megatron adapter of \fmoe is used for MoE structure.
For each layer, $96$ experts are distributed across the GPUs, i.e. $12$ experts are placed on each GPU.
For each input token, the top $2$ experts with highest score are used to process it.
The $d_h$ in expert MLP layer is halved so that the valid FLOPs of the model are almost identical, except for the extra FLOPs introduced by the gate, which is negligible.
Both the baseline model and the MoE model  are trained for $70$ hours.
The \texttt{lm loss} metric in training indicates the convergence tendency of the models.

\begin{figure}[ht]
    \includegraphics[width=\textwidth]{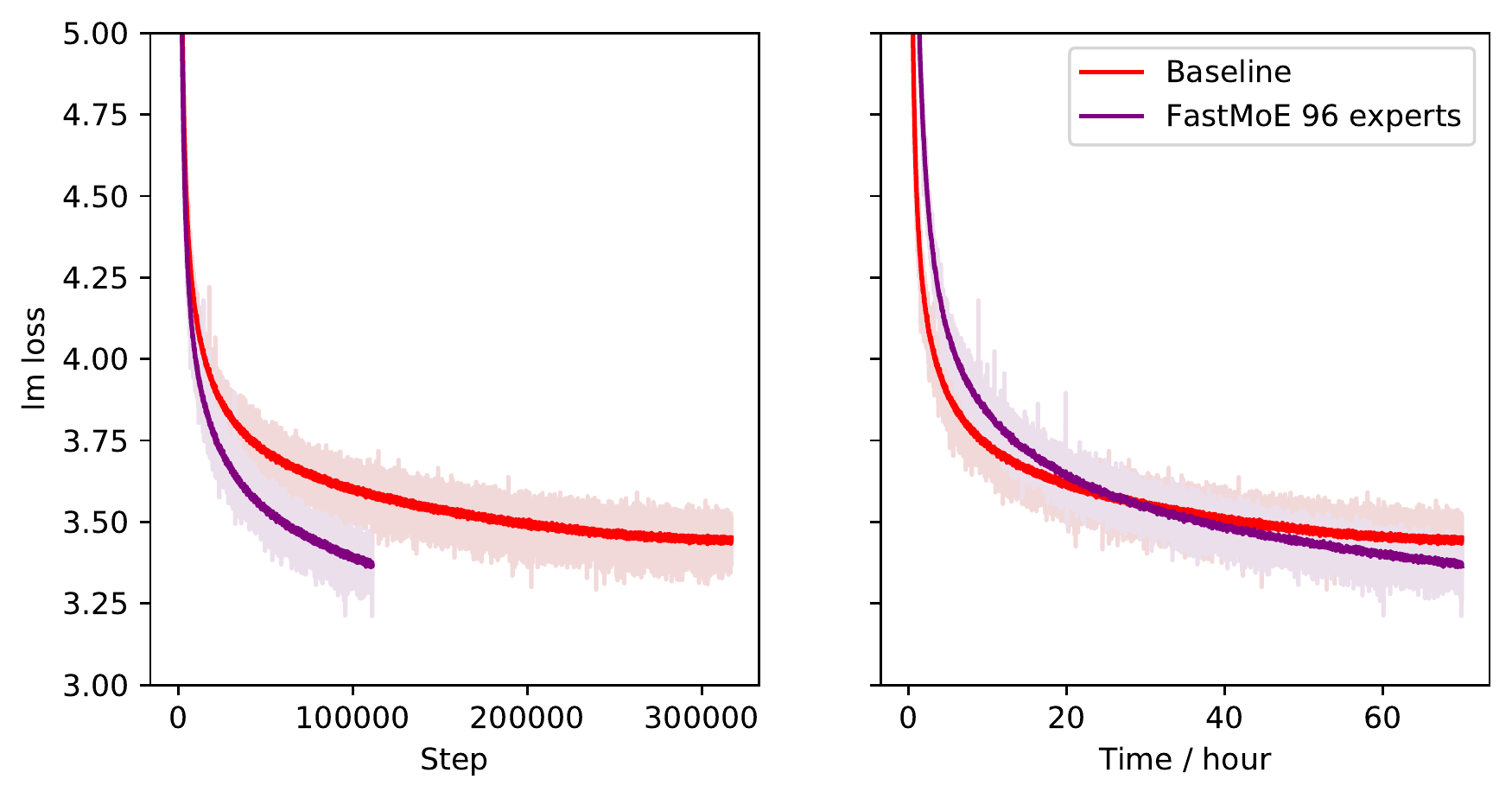}
    \footnotesize The narrow dark lines are smoothed exponentially by $0.97$ from the original loss curve, represented by the brighter wide curves respectively.
    \caption{Loss curve of training a GPT model by \fmoe}
    \label{fig:loss-gpt}
\end{figure}

From \Figref{fig:loss-gpt}, we observed that the training speed of the baseline model is about $3\times$ of \fmoe.
As \fmoe performs more computation and communication, it is a reasonable slow down.
Fortunately, the MoE model achieves much lower loss with the same training iterations.
Also, as a benefit from the efficiency of \fmoe, the MoE model achieves lower loss in the same training time.

\section{Summary and Future Work}
\label{sec:sum}

In this paper, we present \fmoe, an open-source system to train Mixture-of-Experts models.
The system is based on the popular PyTorch framework, and currently supports efficient training on GPUs.
Friendly interfaces of multiple levels are provided for different users to explore different aspects of the MoE architecture.
The performance of \fmoe on a single GPU is well-optimized to exploit the power of GPUs.
\fmoe can also run across GPUs on multiple nodes with reasonable scalability, enabling further enlarging model size.
Real model performance advantage is observed in our end-to-end model training experiment using \fmoe.

We are still working on \fmoe for more features and faster training.
Compared to the GShard model~\citep{chen2020gshard}, \fmoe lacks functionalities to support load-balancing among experts.
The work of load-balance monitor and support for load-balance loss is in progress.
We are also trying to make the system more user-friendly on utilities, such as loading and saving of MoE models.
The performance across multiple GPUs requires joint efforts from the view of both high-performance computing and machine learning.
Any contributions to the open-source project will be appreciated.
We are looking forward to your participation.



\newpage
\bibliography{wpref}
\bibliographystyle{iclr2021_conference}

\end{document}